%% file: main.tex
\documentclass[10pt,twocolumn,letterpaper]{article}

\usepackage[pagenumbers]{cvpr} 


\definecolor{cvprblue}{rgb}{0.21,0.49,0.74}
\usepackage[pagebackref,breaklinks,colorlinks,allcolors=cvprblue]{hyperref}
\usepackage{placeins} 

\def\paperID{000001} 
\def\confName{CVPR}
\def\confYear{2026}

\title{Implicit neural representation of textures}

\author{Albert Kwok$^\dagger$, Zheyuan Hu$^\dagger$, Dounia Hammou \\
    Department of Computer Science and Technology, \\ 
    University of Cambridge, Cambridge, UK. \\
    {\tt\small \{ak2441, zh369, dh706\}@cam.ac.uk}  \\
    {\small $^\dagger$ denotes equal contribution.} 
}

\usepackage{hyperref}
\usepackage{pgffor}
\hypersetup{
    colorlinks=true,
    linkcolor=blue,
    filecolor=magenta,      
    urlcolor=cyan,
}

\begin{document}

\twocolumn[{%
\renewcommand\twocolumn[1][]{#1}%
    \vspace{-14mm}
    \maketitle
    \vspace{-10mm}
    \centering
    \includegraphics[width=0.6\textwidth]{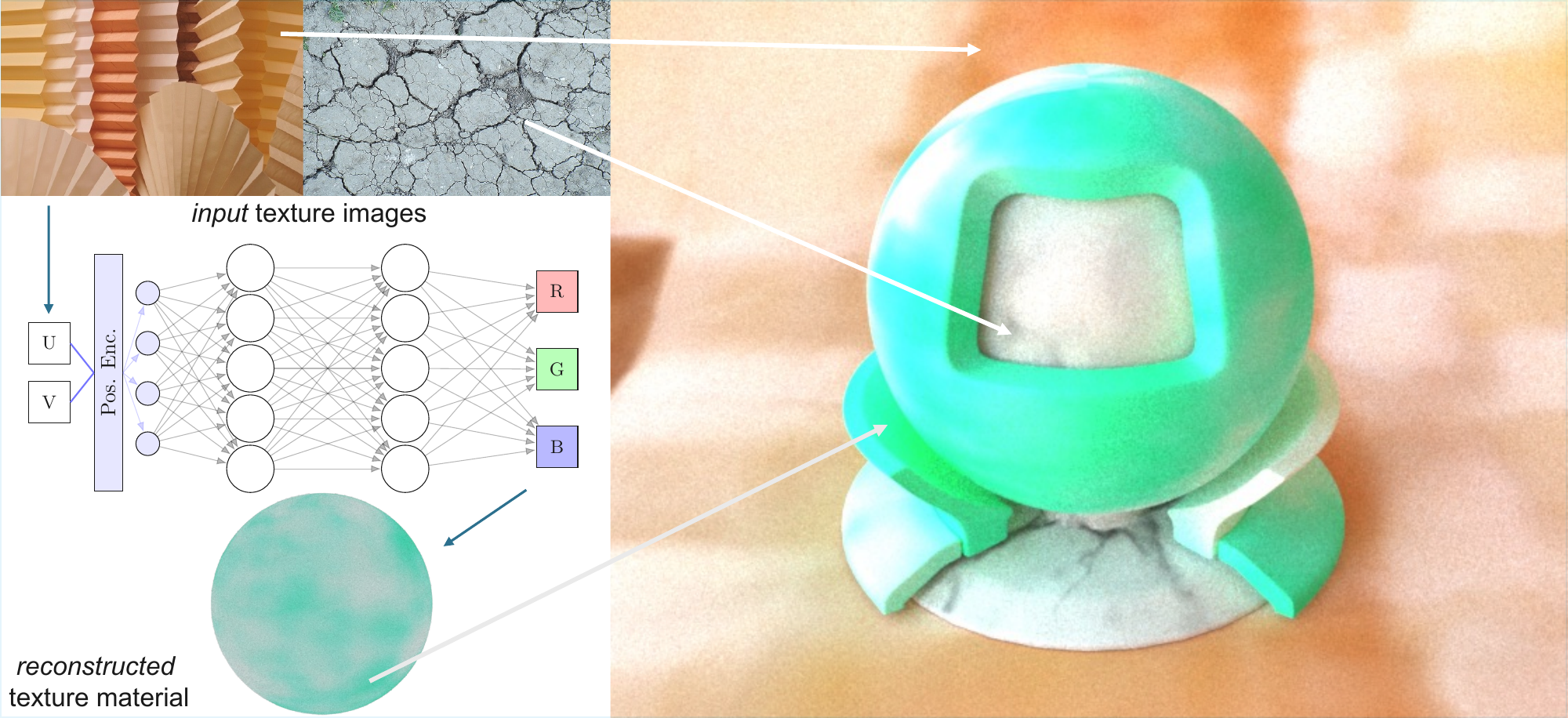}
    \hfill
    \includegraphics[width=0.38\textwidth]{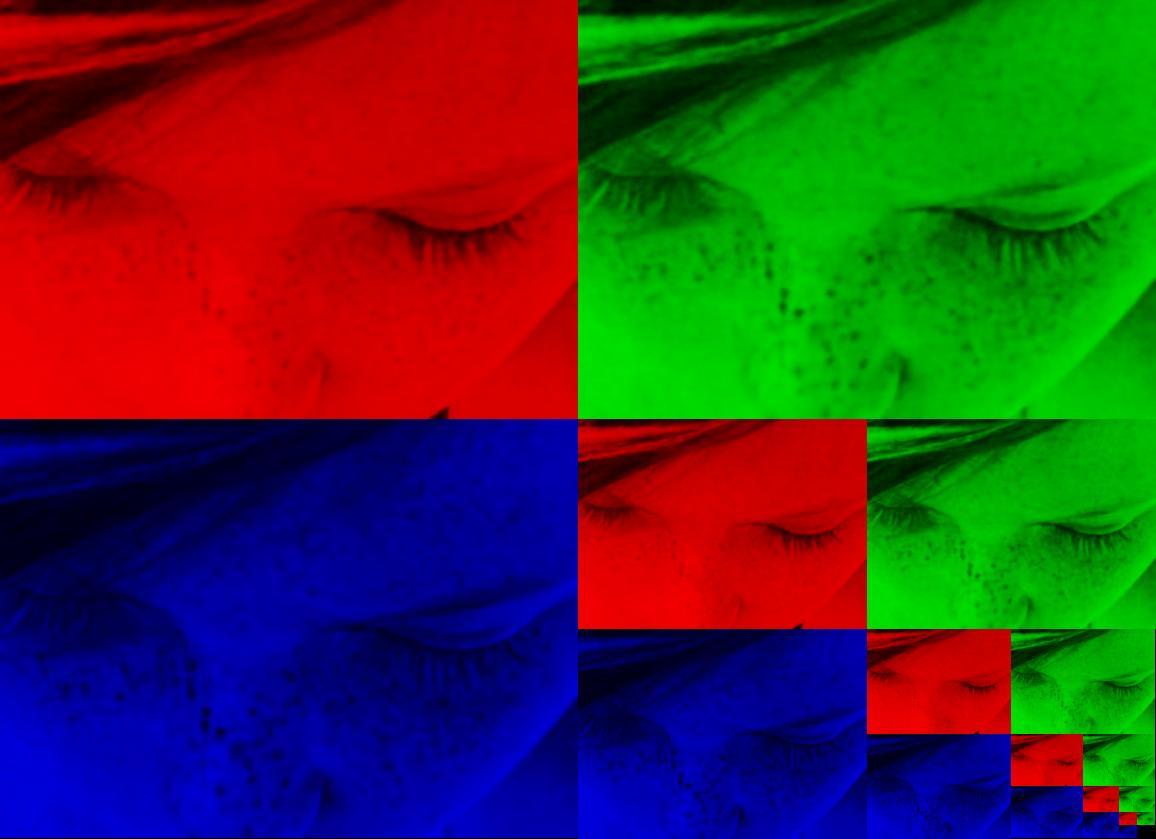}
    \captionof{figure}{(left:) A 3D scene rendered with our INR textures, showcasing high visual quality and straightforward deployment. (right:) A three-channel mipmap pyramid represented by an MLP. \vspace{1em}}
    \label{fig:teaser}
}] 


\input{sec/0_abstract}    
\input{sec/1_intro}
\input{sec/2_related_work}

\input{sec/3_method}

\input{sec/4_evaluation}
\input{sec/5_conclusions}

\clearpage     
{
    \small
    \bibliographystyle{ieeenat_fullname}
    \bibliography{main}
}

\input{sec/X_suppl}

\FloatBarrier

{
    \section{Proposal}
    \input{bids/project_INR_bid}
}
\end{document}

%% file: sec/0_abstract.tex
\begin{abstract}
 %
 Implicit neural representation (INR) has proven to be accurate and efficient in various domains. In this work, we explore how different neural networks can be designed as a new texture INR, which operates in a continuous manner rather than a discrete one over the input UV coordinate space. Through thorough experiments, we demonstrate that these INRs perform well in terms of image quality, with considerable memory usage and rendering inference time. We analyze the balance between these objectives. In addition, we investigate various related applications in real-time rendering and down-stream tasks, e.g. mipmap fitting and INR-space generation.
 %
 %
 \vspace{-4mm}
\end{abstract}

%% file: sec/1_intro.tex
\section{Introduction}
\label{sec:intro}



Implicit Neural Representations (INRs) are effective by using neural networks to directly represent continuous, coordinate-based signals (such as images \cite{Strumpler2022INRImg}, shapes, audio, or physical fields) rather than discrete grids or explicit parametrizations \cite{Sitzmann2020INRActivation}. Examples are multilayer perceptron (MLP) and convolutional neural network (CNN).


Textures are one of the largest GPU memory usage and account for the majority of the energy consumed by processors \cite{Silpa08TexMemoryUsage}. Thus, INR presents a memory- and power-efficient method, while maintaining high rendering quality. Furthermore, the compact form also helps with downstream tasks, including material generation and asset baking.

A tiny multilayer perceptron (MLP) or other instant neural models designed for graphics primitives will take as input the UV coordinates and output the corresponding RGB color. The optimized network weights will form a compressed representation of textures.

The contributions of our work are as follows, (1) implement four different INRs of textures, evaluating three of them in terms of performance, efficiency and memory usage; (2) integrate INRs into Mitsuba 3 \cite{mitsuba3}\footnote{Mitsuba 3 is a customizable Python / C++ renderer.}, by extending the plugin that reconstructs from MLP weights during rendering, as shown in Figure \ref{fig:renderer-sphere}; (3) explore downstream tasks, including mipmap fitting and INR weight-space generation.


%% file: sec/2_related_work.tex
\section{Related work}
\label{sec:related-work}



\paragraph{Texture compression.} Inspired by discrete cosine transform (DCT), JPEG \cite{JPEG} is an texture image compression method. Adaptive Scalable Texture Compression (ASTC) \cite{Nystad2012ASTC} provides lossy block-based compression over a range of bit rates, i.e. 1 to 8 bits per pixel (bpp).  In this work, we present neural network as an alternative method with good compression ratio and content-aware adaptation. 

\paragraph{INR design.} Neural network, e.g. Multi-Layer Perceptron (MLP), is a universal function approximator \cite{HORNIK1989MLPUniversal} for continuous functions. Given discrete samples of data, prior work \cite{Strumpler2022INRImg} overfits the MLP, which becomes an implicit neural representation (INR) of images. Below is an overview of different design choices and our contributions.

\textbf{Periodic activation functions} have been proposed in MLP \cite{Sitzmann2020INRActivation}, which have been proven to be an efficient signal representation. 

\textbf{Positional encoding} is crucial to leveraging input from low dimension to high dimension for efficient training. Examples are Fourier encoding \cite{Tancik2020FourierEncoding} and multiresolution hash encoding \cite{Muller2022HashEncoding}. The former has been applied in image INR \cite{Strumpler2022INRImg} via MLP. 


In this work, we examine the effectiveness and potential limitations of periodic activation functions and explore both positional encoding schemes in the texture INR domain.

%% file: sec/3_method.tex
\section{Method}
\label{sec:method}



\subsection{Dataset analysis and sample selection} 

\paragraph{Texture dataset.} Describable Textures Dataset (DTD) \cite{cimpoi14DTDTex} consists of 47 different categories from human perception and a total of 5640 images. From which, we perform analysis and select the most diverse samples based on metrics. 



To filter out the most diverse images $I(u, v)$ in the frequency domain, we utilize the \textbf{Lap}lacian \textbf{V}ariance (LAPV) \cite{memon2016imagequality}, also known as Laplacian response, as a measure of focus or sharpness.  While different image complexity metrics exist, e.g. compression ratio, image entropy, and edge density, we use LAPV due to its correlation with perceived quality, i.e. $\text{FocusMeasure} := \text{LAPV}(I) = \text{Var}(\Delta I(u,v))$.

\begin{figure}[htbp]
    \centering
    \includegraphics[width=0.8\linewidth]{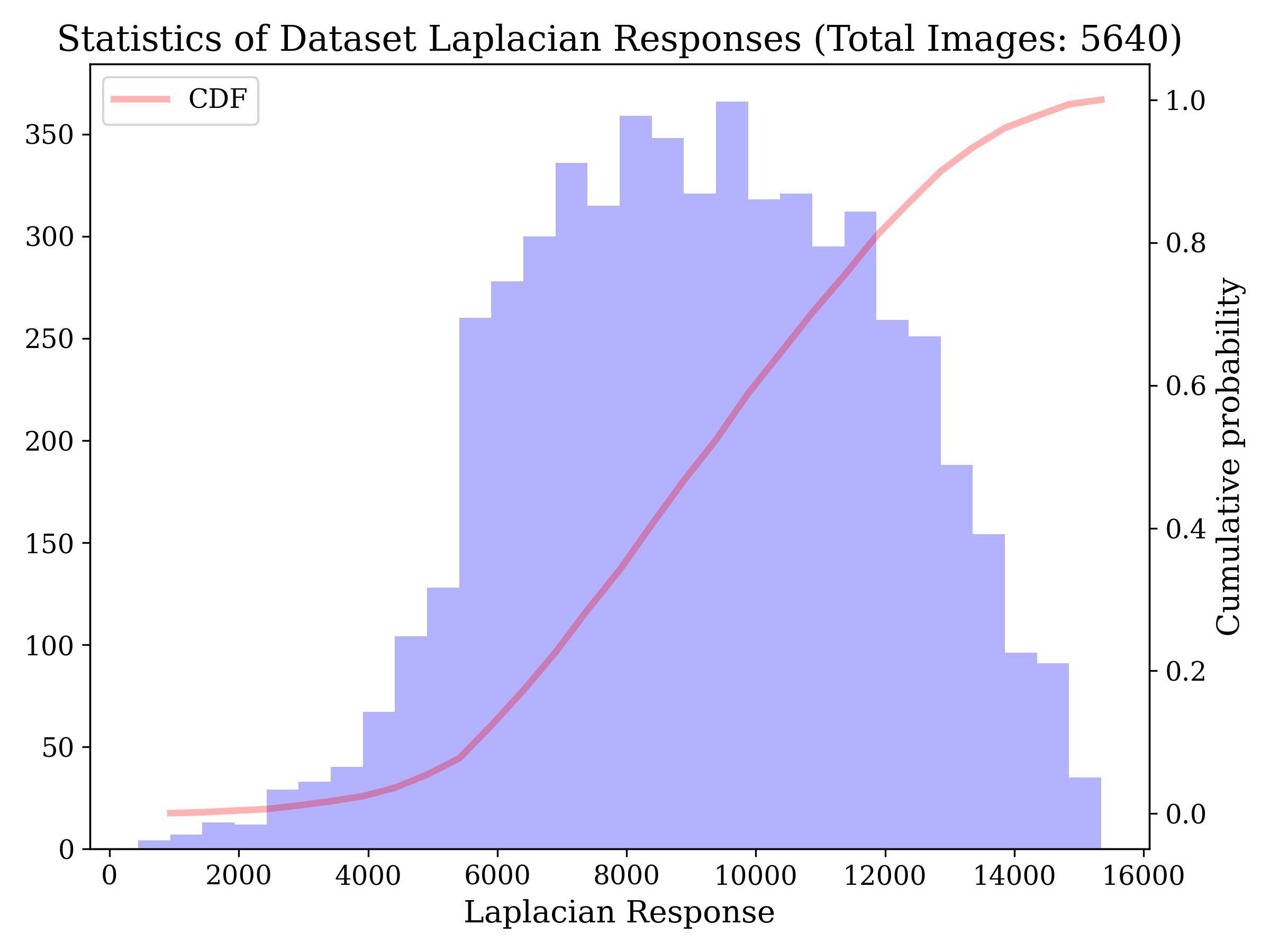}
    \caption{Laplacian response histogram with its Cumulative Distribution Function (red curve).}
    \label{fig:LAPV-hist}
\end{figure}

The dataset Laplacian response histogram is shown in Figure \ref{fig:LAPV-hist}, from which we sampled 25 images (Figure \ref{fig:sampled-textures}) at regular intervals.


\subsection{INR architectures}
\paragraph{Multi-Layer Perceptron (MLP)} \cite{Popescu2009MLP}, also known as Feed-forward Neural Network (FNN), consists of input, hidden and output layers with dimensions $d_\text{in}$, $d_\text{h}$ and $d_\text{out}$. Each layer contains several neural nodes, which apply a non-linear activation function $f_\text{act}$ (e.g. ReLU \cite{agarap2019ReLU}) after the linear combination of values $\textbf{a}^{l-1} \in \mathbb{R}^{d_\text{in}}$ from previous $(l-1)$-th layer. The $l$-th layer output is thus,
$$
\mathbf{a}^l = f_\text{act}(\textbf{W}^{l} \textbf{a}^{l-1} + \textbf{b}^l),
$$
where weight matrix is $\textbf{W}^l \in \mathbb{R}^{d_\text{in} \times d_{\text{out}}}$ and bias term $\textbf{b}^l \in \mathbb{R}^{d_\text{out}}$.\footnote{Here we assume no hidden layers exist. Otherwise, use $d_h$ instead of $d_\text{in}$ when $l \geq 2$ or $d_\text{out}$ when $l=1$.}


\textbf{SIREN} \cite{Sitzmann2020INRActivation} is a special type of MLP, with sinusoidal activation functions rather than ReLU, i.e. $f_\text{act}=\sin(\cdot)$, which increases overfitting for images, video, sound, and their derivatives, even without any positional encoding. 

We observe that SIREN is sensitive to the learning rate $\eta$, leading to distinct converged results. Hence, a smaller $\eta$ was adopted. Its network weights $\textbf{W}$ are distinctly initialized via a uniform distributions defined as below,
$$
\begin{aligned}
\text{First layer:} & \quad
W \sim \mathcal{U}\Bigg(-\frac{1}{d_{\text{in}}}, \frac{1}{d_{\text{in}}}\Bigg), \\
\text{Hidden layers:} & \quad
W \sim \mathcal{U}\Bigg(-\frac{\sqrt{6 / d_{\text{in}}}}{\omega_0}, \frac{\sqrt{6 / d_{\text{in}}}}{\omega_0}\Bigg),
\end{aligned}
$$
where $d_\text{in}$ is the input dimension of the layer and sine frequency scaling $\omega_0$ is the hyperparameter.




\textbf{Fourier encoding.} To learn high-frequency functions effectively, we utilize the Fourier feature mapping \cite{Tancik2020FourierEncoding} as positional encoding. By concatenating the input UV coordinates $\textbf{v}$ and their frequency features, the input to MLP is thus,
$$
\text{Pos}(\textbf{v}) = [\textbf{v}, \sin(2 \pi f_i \textbf{v})_{i=1}^{n_f}, \cos(2\pi f_i \textbf{v})_{i=1}^{n_f}],
$$
where $n_f$ is the number of frequency bands and index $i$ loops over all frequencies. They are treated as hyperparameters, adjusted based on validation results.

We also implement the extension of using multi-resolution hash encoding \cite{Muller2022HashEncoding}. However, due to the limited texture resolution, we believe that this would offer little added performance benefit. The assessment of this approach on higher-resolution images is deferred to future work.
 
\begin{figure*}[htbp]
    \centering
    \begin{subfigure}[b]{0.4\textwidth}
        \centering
        \includegraphics[width=\textwidth]{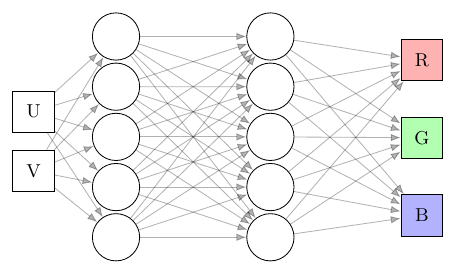}
        \caption{General MLP architecture without extra encoding.}
        \label{fig:mlp}
    \end{subfigure}
    \hfill
    \begin{subfigure}[b]{0.45\textwidth}
        \centering
        \includegraphics[width=\textwidth]{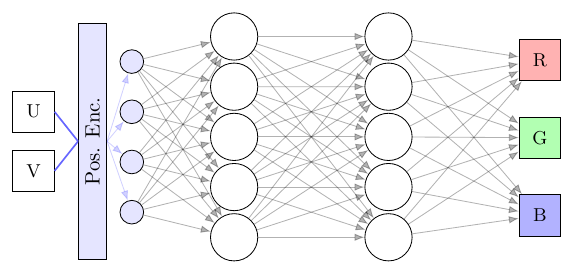}
        \caption{General MLP architecture with positional encoding.}
        \label{fig:mlp_pe}
    \end{subfigure}
    
    \caption{MLP architectures used for INR texture, both without and with positional encoding (pos. enc.).}
    \label{fig:mlp_comparison}
\end{figure*}


In conclusion, we implement \textbf{four} different MLP architectures, as shown in Figure \ref{fig:mlp_comparison}, and evaluate \textbf{three} of them, i.e., (1) a pure MLP with no positional encoding, (2) SIREN, and (3) an MLP with fourier positional encoding.

In order to vary the size of the model to compare the performance against the bitrate, we varied the width and depth of the models. For each architecture, we tested the models with 1, 2, or 3 hidden layers of sizes 128, 256, and 512, (except for 3 layers of 512). All of the models were tested with the optimisers Adam and Rprop. Adam generally produced better results across the board. 


\subsection{Mipmap fitting}

To model mipmaps, instead of directly integrating over MLPs $: (u,v) \rightarrow \mathbb{R}^3$, the models were given an additional normalized input parameter $t \in [0,1]$, which denoted the level of detail. In this work, we tested models against 6 mipmap levels. The retraining on the augmented dataset removes the hallucinations at points the network hasn’t explicitly seen during training. 


The mipmaps were generated by downscaling the base image by bilinear downsampling\footnote{We use \texttt{resize($\cdot$)} from \texttt{scikit-image} with a default order of 1 in spline interpolation, i.e. bilinear downsampling.}. The generated images were then paired with their relevant $u$, $v$, and $t$ coordinates to produce the final dataset. In this way, each pixel sample is equally weighted, such that the higher resolution images have a greater impact on the loss than the lower resolution images. 


\subsection{INR-space generation} \label{sec:main-INR-generation}


Inspired by both image-space and neural weight-space generation \cite{M3ashy2026}, we investigate generation on the new data modality of textures, i.e. INR. The goal is to generate new MLP weights, which is equivalent to novel textures after inference. To achieve this, we deploy generative methods, e.g. diffusion or VAE, on MLP weights\footnote{This feature was implemented at \texttt{INR\textunderscore space\textunderscore generation} branch.}. They are termed hypernetworks \cite{Erkoc_2023_HyperDiffusion}, due to the nature of taking neural network weights as input. It reduces the input dimensionality and thus training cost. The evaluation focuses on the fidelity and coverage between synthetic and reference sets.

\begin{figure}[htbp]
  \centering
\foreach \i in {0,...,5}{%
    \begin{subfigure}{0.16\linewidth}
      \centering
      \includegraphics[width=\linewidth]{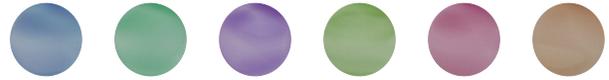}
    \end{subfigure}\hfill
  }
  \caption{Texture RGB augmentation for generation.}
  \label{fig:tex_rbg_augmentation}
\end{figure}

\paragraph{Data augmentation.} To avoid underfit for generation models, we adopt the RGB permutation (Figure \ref{fig:tex_rbg_augmentation}) to increase the size of training dataset. As an extension, linear interpolation of two randomly selected textures could be adopted in future work.


\textbf{Diffusion model} is modelled by the Markov chain along time steps $t \in [0, T] \subset \mathbb{Z}^+$. The forward process $q(x_t | x_{t-1}$ adds Gaussian noise to the input data $x_0$, while backward process $p_{\theta}(x_{t-1}|x_t)$ removes noise from the random Gaussian samples $x_T$. In our work, a separate transformer-based network predicts the $p_{\theta}(x_{t-1}|x_t)$. For mathematical derivation, please refer to \cite{su2025chord, M3ashy2026}.

%% file: sec/4_evaluation.tex
\section{Evaluation}
\label{sec:eval}



\paragraph{Evaluation metrics.} We evaluate the reconstructed texture $I^r$ quality of INR by measuring the pixel-wise errors (MAE, MSE, PSNR) and perceptual and structural similarity (SSIM, LPIPS, VMAF) against the ground truth $I^{gt}$. 


Define the coloured image as $I_{w,h,c}: \{1, ..., W\} \times \{1, ..., H\} \times [1, ..., C] \rightarrow [0, 1]$, where $W$ and $H$ are the  width and height, and $C$ is the number of colour channels. 
\textbf{MAE (Mean Absolute Error)} measures the average absolute pixel differences between $I^r_{w,h}$ and $I^{gt}_{w,h}$,
$$
MAE = \frac{1}{W \times H \times C} \sum_{w=1}^W \sum_{h=1}^H  \sum_{c=1}^C|I^r_{w,h,c} - I^{gt}_{w,h,c}|.
$$
\textbf{MSE (Mean Squared Error)} measures the average squared pixel differences between $I^r_{w,h}$ and $I^{gt}_{w,h}$,
$$
MSE = \frac{1}{W \times H \times C} \sum_{w=1}^W \sum_{h=1}^H \sum_{c=1}^C |I^r_{w,h,c} - I^{gt}_{w,h,c}|^2.
$$
\textbf{PSNR (Peak Signal-to-Noise Ratio)} \cite{Hore2010PSNR} is a normalized fraction of the maximum signal power to that of the noise, where $MAX_I \in [0, 1]$ is the maximum pixel value of images,
$$
PSNR = 10 \log_{10} \frac{MAX_I}{MSE} [dB].
$$
\textbf{SSIM (Structural Similarity Index)} \cite{Wang2004SSIM}, as a perception-based metric, approximates local luminance $mu_I$ , contrast $\sigma_I$ and structure $\sigma_{I^r, I^{gt}}$ via the image I’s Y-channel $I^Y$ mean, variance and covariance respectively,
$$
\mu_I = \bar{I^Y}, \quad \sigma_I = Var(I^Y), \quad \sigma_{r, gt} = Cov(I^r, I^{gt}).
$$
Luminance, contrast, and structure are defined below, 
$$
\begin{aligned}
& l = \frac{2\mu_{I^r}\mu_{I^{gt}} + C_1}{\mu_{I^r}^2 + \mu_{I^{gt}}^2 + C_1}, \quad
c = \frac{2\sigma_{I^r}\sigma_{I^{gt}} + C_2}{\sigma_{I^r}^2 + \sigma_{I^{gt}}^2 + C_2}, \\
& s = \frac{\sigma_{I^r, I^{gt}} + C_3}{\sigma_{I^r}\sigma_{I^{gt}} + C_3}, \quad \text{constants } C_1, C_2, C_3 \in \mathbb{R}.
\end{aligned}
$$
SSIM is defined by the product of the luminance, contrast and structure functions above,
$$
SSIM= l^\alpha \cdot c^\beta \cdot s^\gamma, \quad \alpha, \beta, \gamma > 0.
$$
\textbf{LPIPS (Learned Perceptual Image Patch Similarity)} \cite{zhang2018perceptual} extracts perceptional features from deep learning networks, where $f_l$ are the feature maps at layer $l$ with corresponding weights $w_l$,
$$
LPIPS = \sum_l \frac{1}{W \times H} \sum_{w, h} || w_l \odot ||f_l(I^r) - f_l (I^{gt})||_2^2.
$$
\textbf{VMAF (Video Multimethod Assessment Fusion)} \cite{li2016vmaf}, originally developed by Netflix, applies a SVM regression on image perception features VIF (visual information fidelity), ADM (Additive Detail Metric),  and the motion feature. Here, we only applied the metric to images rather a video, i.e. setting number of frames to be 1. For details, please refer to \cite{vmaf-torch}.

 

\subsection{Single texture performance evaluation}

\begin{figure*}[htbp]
  \centering
  \foreach \img/\cap/\lab in {
      per_arch_lpips_bucket_bpp.png/Performance measured by LPIPS./core_lpips,
      per_arch_vmaf_bucket_bpp.png/Performance measured by VMAF./core_vmaf,
      per_arch_ssim_bucket_bpp.png/Performance measured by SSIM./core_ssim,
      per_arch_psnr_bucket_bpp.png/Performance measured by PSNR./core_psnr
  }{%
    \begin{subfigure}{0.45\textwidth}
      \centering
      \includegraphics[width=\linewidth]{figs/\img}
      \caption{\cap} \label{fig:\lab}
    \end{subfigure}
  }
  \caption{Texture INR performance against bitrate (bits per pixel).}
  \label{fig:core_evals}
\end{figure*}


Figure \ref{fig:core_evals} shows the plots for the performance for the INRs against the bitrate of the model, for each model architecture. To reduce noise, this plot is bucketed by rounding the bitrate to the nearest 2 bits. This was done because the images in the dataset have differing sizes, and therefore the results have a large variety of bitrates, which produces visual noise when plotted. For LPIPS, both the SIRENs and the Fourier MLPs performed very well, being very close to the maximum similarity to the original of 0, though the SIRENs had a noticeable drop in performance for very high compression. On the other hand, the pure MLP clearly struggles to meet the performance of the other two. This shows that the INRs effectively captured key features of the image, effectively capturing the high-level structure of the image. 

The SSIM, VMAF, and PSNR of the INRs are more mixed, with a noticeable difference to the original. Across both metrics, the best performing INRs are the Fourier MLPs, followed by the SIRENs. All of the models have broadly similar curves, implying that the models all struggle to accurately learn similar features. These results imply that the INRs may struggle to accurately capture details. This could be improved by adjusting the frequency parameters of the Fourier MLPs and the SIRENs. 


\subsection{Qualitative architecture evaluation}

\label{sec:qualitative eval}

\begin{figure}
    \centering
    \includegraphics[width=0.5\linewidth]{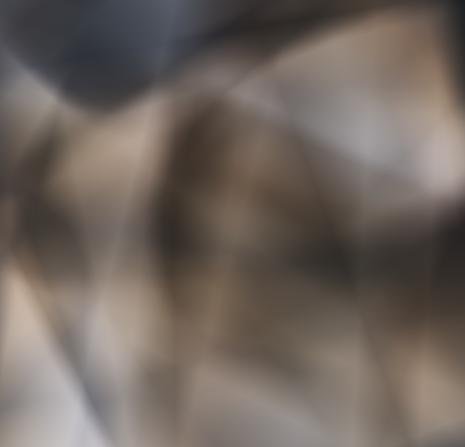}
    \includegraphics[width=0.48\linewidth]{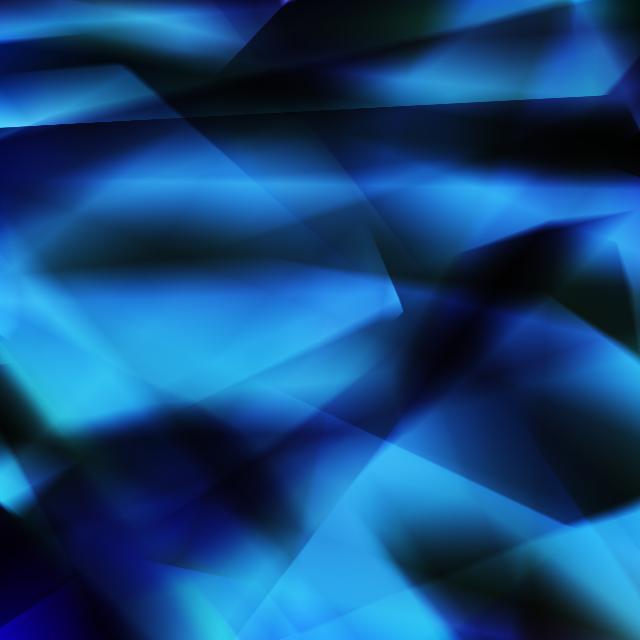}
    \caption{Some artefacts learned by the pure MLPs.}
    \label{fig:mlp artefacts}
\end{figure}

\begin{figure}
    \centering
    \includegraphics[width=0.51\linewidth]{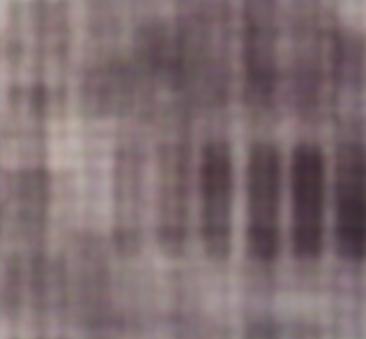}
    \includegraphics[width=0.47\linewidth]{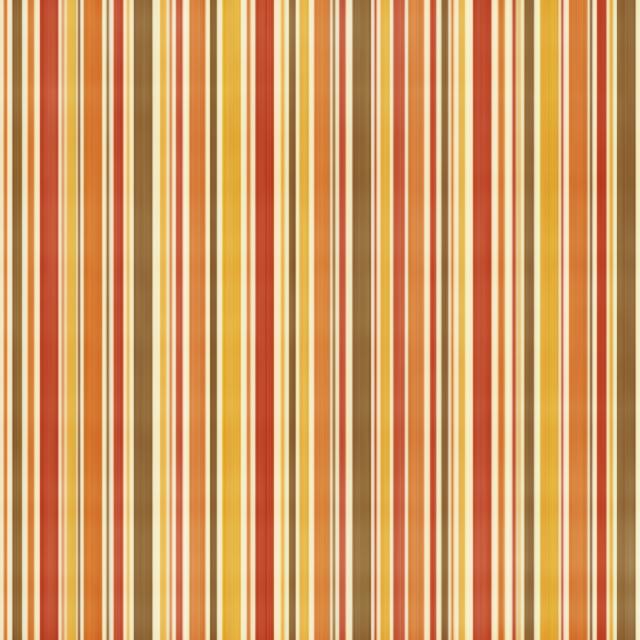}
    \caption{Some artefacts learned by the fourier encoded MLPs.}
    \label{fig:fourier artefacts}
\end{figure}

\begin{figure}
    \centering
    \includegraphics[width=0.385\linewidth]{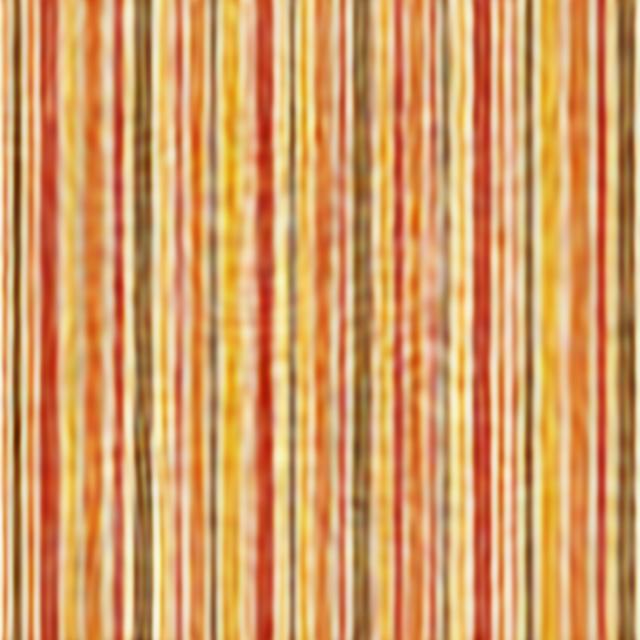}
    \includegraphics[width=0.58\linewidth]{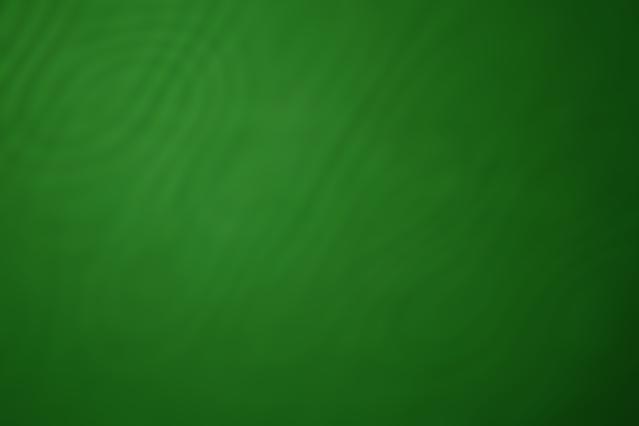}
    \caption{Some artefacts learned by the SIRENs.}
    \label{fig:sine artefacts}
\end{figure}

The different architectures produced different kinds of artefacts. 

The simplest are those of the pure MLP, which consistently produced blurry results as expected: Fourier encoding and SIRENs were proposed to help MLPs learn higher frequency features. These blurry results sometimes had clear, straight lines through them, which if the image consisted mostly of straight lines allowed the pure MLP to achieve reasonable results. Some of these effects are shown in Figure \ref{fig:mlp artefacts}.

The Fourier encodings usually produced straight line artefacts, especially vertical and horizontal lines, reflective of the structure of the encoding. These artefacts were usually successfully trained out, but can be seen in early epochs. After training, these models still frequency produce a grainy noise, which degrades its ability to learn large areas of flat colours. This noise is perceptually weak, whilst still having significant per pixel impacts. Some of these effects are shown in Figure \ref{fig:fourier artefacts}.

The SIRENs produced lumpy curve artefacts. These were harder to overcome, and can be still clearly seen affecting the results of several images. This may reflect why the SIRENs sometimes struggled to match performance with the other models: these artefacts can mean it struggles to learn straight lines, with noticeably poor performance on very geometric textures, as shown by Figure \ref{fig:sine artefacts}.

\subsection{Comparison of Adam and Rprop}

\begin{figure*}[htbp]
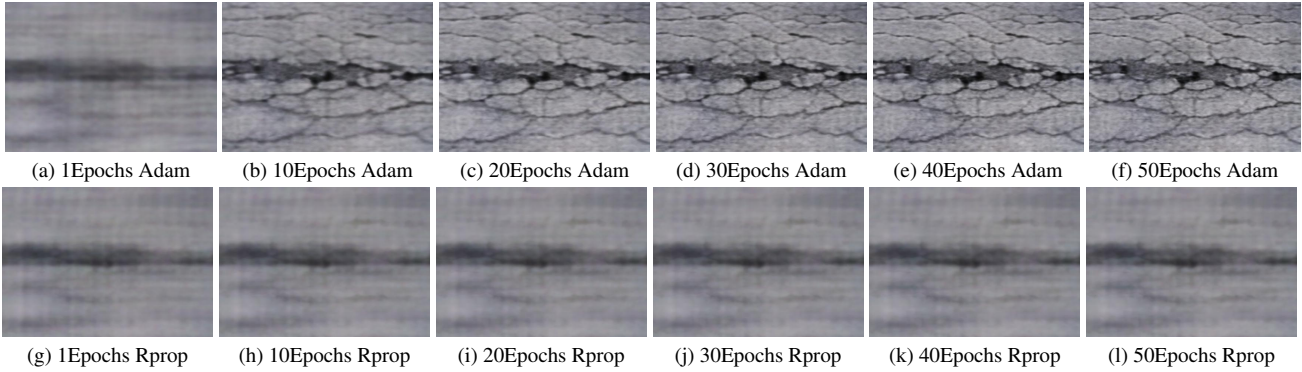

  \centering
  \foreach \epoch/\lab in {
      1/1,
      10/10,
      20/20,
      30/30,
      40/40,
      50/50
  }{%
    \begin{subfigure}{0.16\textwidth}
      \centering
      \includegraphics[width=\linewidth]{figs/learning_snapshots/uv_fourier_mlp_128x2_potholed_0113_\epoch_1e-3_adam.jpg}
      \caption{\epoch Epochs Adam} \label{fig:fourier adam \lab}
    \end{subfigure}
  }
  \foreach \epoch/\lab in {
      1/1,
      10/10,
      20/20,
      30/30,
      40/40,
      50/50
  }{%
    \begin{subfigure}{0.16\textwidth}
      \centering
      \includegraphics[width=\linewidth]{figs/learning_snapshots/uv_fourier_mlp_128x2_potholed_0113_\epoch_1e-3_rprop.jpg}
      \caption{\epoch Epochs Rprop} \label{fig:fourier rprop \lab}
    \end{subfigure}
  }
  \caption{The texture learned by a Fourier encoded MLP with 128$\times$2 hidden layers learning with Adam and Rprop.}
  \label{fig:fourier adam v rprop}
\end{figure*}

\begin{figure*}[htbp]
  \centering
  \foreach \epoch/\lab in {
      1/1,
      10/10,
      20/20,
      30/30,
      40/40,
      50/50
  }{%
    \begin{subfigure}{0.16\textwidth}
      \centering
      \includegraphics[width=\linewidth]{figs/learning_snapshots/uv_sine_mlp_256x3_crosshatched_0164_\epoch_1e-4_adam.jpg}
      \caption{\epoch Epochs Adam} \label{fig:sine adam \lab}
    \end{subfigure}
  }
  \foreach \epoch/\lab in {
      1/1,
      10/10,
      20/20,
      30/30,
      40/40,
      50/50
  }{%
    \begin{subfigure}{0.16\textwidth}
      \centering
      \includegraphics[width=\linewidth]{figs/learning_snapshots/uv_sine_mlp_256x3_crosshatched_0164_\epoch_1e-4_rprop.jpg}
      \caption{\epoch Epochs Rprop} \label{fig:sine rprop \lab}
    \end{subfigure}
  }
  \caption{The texture learned by a SIREN with 256$\times$3 hidden layers learning with Adam and Rprop.}
  \label{fig:sine adam v rprop}
\end{figure*}

\begin{figure*}[htbp]
  \centering
  \foreach \epoch/\lab in {
      1/1,
      10/10,
      20/20,
      30/30,
      40/40,
      50/50
  }{%
    \begin{subfigure}{0.16\textwidth}
      \centering
      \includegraphics[width=\linewidth]{figs/learning_snapshots/uv_mlp_256x3_bumpy_0101_\epoch_1e-3_adam.jpg}
      \caption{\epoch Epochs Adam} \label{fig:mlp adam \lab}
    \end{subfigure}
  }
  \foreach \epoch/\lab in {
      1/1,
      10/10,
      20/20,
      30/30,
      40/40,
      50/50
  }{%
    \begin{subfigure}{0.16\textwidth}
      \centering
      \includegraphics[width=\linewidth]{figs/learning_snapshots/uv_mlp_256x3_bumpy_0101_\epoch_1e-3_rprop.jpg}
      \caption{\epoch Epochs Rprop} \label{fig:mlp rprop \lab}
    \end{subfigure}
  }
  \caption{The texture learned by a simple MLP with 256$\times$3 hidden layers learning with Adam and Rprop.}
  \label{fig:mlp adam v rprop}
\end{figure*}

Overall, Adam produced notably better and more consistent results than Rprop. The difference can be seen visually in the learning snapshots presented in Figures \ref{fig:fourier adam v rprop}, \ref{fig:sine adam v rprop}, \& \ref{fig:mlp adam v rprop}. Across all three examples, the Rprop is clearly blurrier, with some of the images having additional artefacts from the learning process, such as the rectangular structures as a result of the high frequencies of the Fourier encoding. 

It is worth noting that the given image for the pure MLPs represents an unusually large difference in performance between Adam and Rprop: for most of the images learned by the pure MLP neither optimiser made significant progress to accurately learning the image. 

\textbf{Conclusion} Adam has demonstrated itself to be the better optimiser across the board. Whilst it is possible that hyper parameter tuning could improve Rprop's performance, we doubt that it will improve performance enough to consistently and significantly outperform Adam. 

\subsection{Comparison to ASTC} \label{sec:comp-astc}

Figure \ref{fig:comp_evals} compares the performance of the best architecture (Fourier MLPs) against ASTC. As the graph clearly demonstrates, whilst ASTC achieves good compression ratios, it has a clear sacrifice in quality, especially in LPIPS. The VMAF plot is representative of the results for SSIM and PSNR (which are in Figure \ref{fig:comp_evals_2}): whilst the INR does perform consistently better, the result is noisy, and is less significant than the gain in LPIPS.

\begin{figure*}[htbp]
  \centering
  \foreach \img/\cap/\lab in {
      comp2_inr_lpips_bucket_bpp.png/Performance measured by LPIPS./comp2_inr_lpips,
      comp2_inr_vmaf_bucket_bpp.png/Performance measured by VMAF./comp2_inr_vmaf
  }{%
    \begin{subfigure}{0.45\textwidth}
      \centering
      \includegraphics[width=\linewidth]{figs/\img}
      \caption{\cap} \label{fig:\lab}
    \end{subfigure}
  }
  \caption{ASTC and the best INR performance against bitrate (bits per pixel).}
  \label{fig:comp_evals}
\end{figure*}

\begin{figure*}[htbp]
  \centering
  \foreach \img/\cap/\lab in {
      comp2_inr_ssim_bucket_bpp.png/Performance measured by SSIM./comp2_inr_ssim,
      comp2_inr_psnr_bucket_bpp.png/Performance measured by PSNR./comp2_inr_psnr
  }{%
    \begin{subfigure}{0.45\textwidth}
      \centering
      \includegraphics[width=\linewidth]{figs/\img}
      \caption{\cap} \label{fig:\lab}
    \end{subfigure}
  }
  \caption{ASTC and the best INR performance against bitrate (bits per pixel).}
  \label{fig:comp_evals_2}
\end{figure*}

\paragraph{Efficiency.} Training occurred at around 0.5-2 iterations per second on a device with an RTX 5080Ti, for a total of 50 to 200 seconds for all 50 iterations. This makes it practical to use in non-real-time applications, though is slow enough that extensive hyper-parameter search is still a large time investment. 

Rendering time with MLP-based texture inference scales proportionally with the total number of samples per pixel (SPP) and the aggregated seeds. For $spp=1$ and $seed=1$, the teaser rendering time on an Apple M1 CPU is 4.7 seconds.

\subsection{Mipmapped texture performance evaluation}

\begin{figure}
    \centering
    \includegraphics[width=0.54\linewidth]{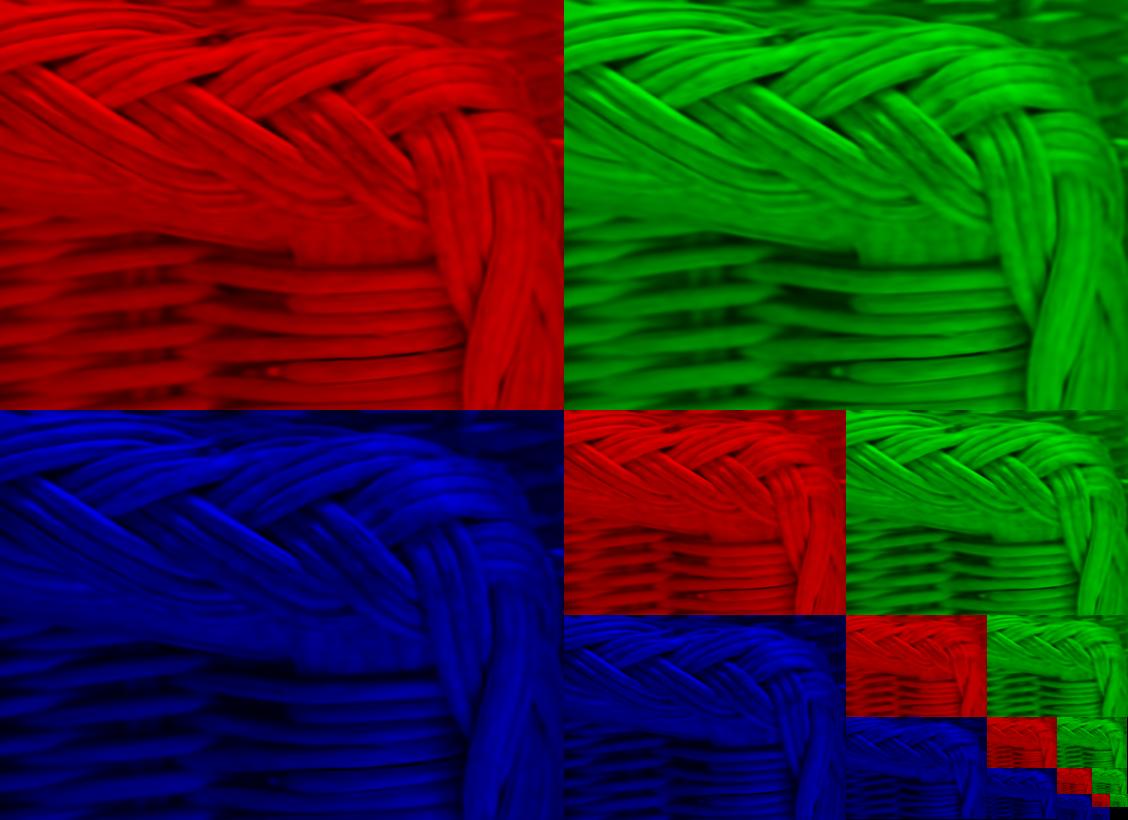}
    \includegraphics[width=0.337\linewidth]{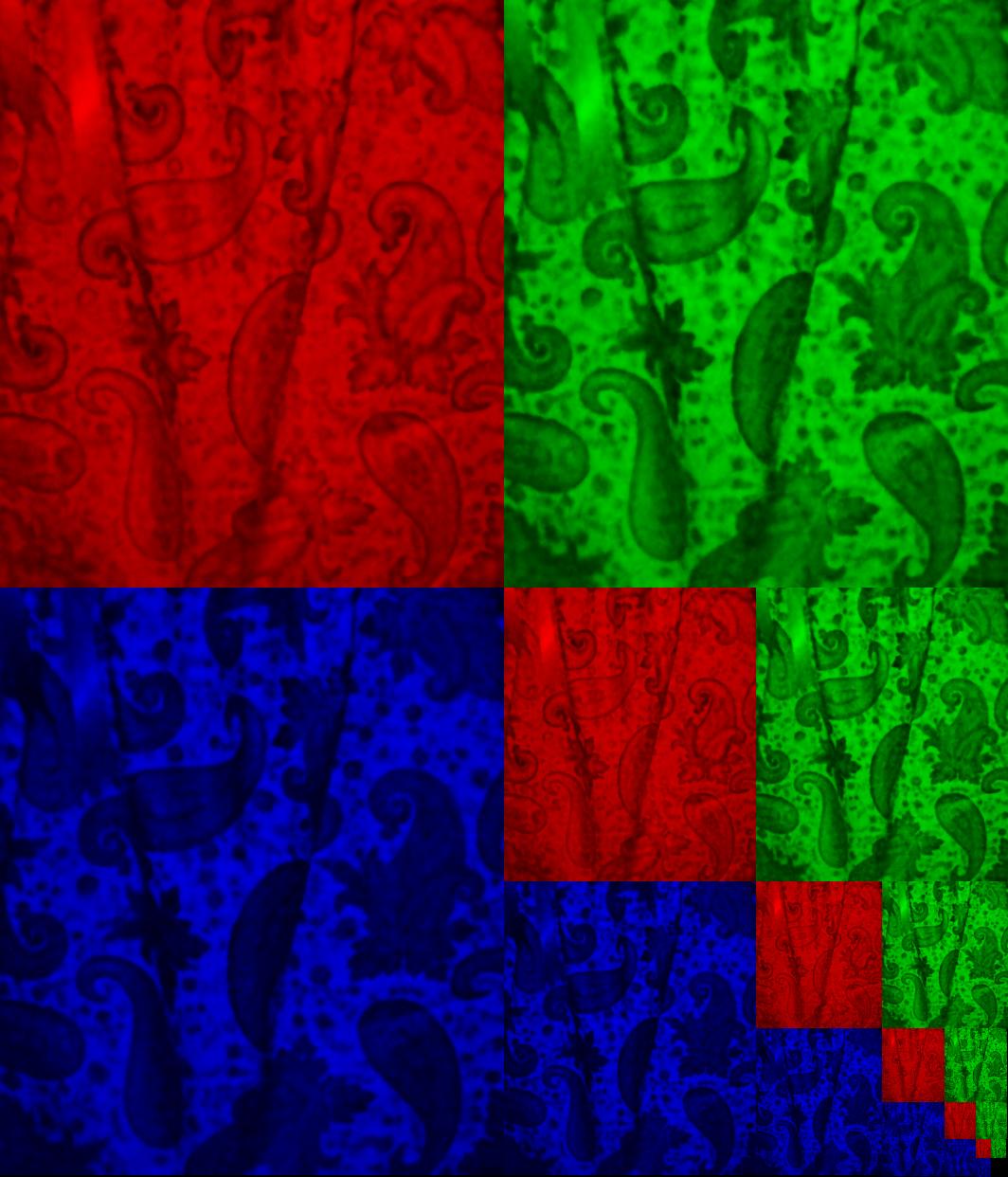}
    \caption{Some learned mipmap pyramids, with the figure on the left being modelled by a SIREN, and the figure on the right being modeeled by a Fourier encoded MLP}
    \label{fig:mipmap sample}
\end{figure}


Figure \ref{fig:mipmap_6_evals} shows that the models have model mipmapped textures well, with similar results to textures without mipmapping. This demonstrates that adding mipmapping has little impact to the performance of the model. Surprisingly, the VMAF was significantly better than without mipmapping. This is visually confirmed by Figure \ref{fig:mipmap sample}.

\begin{figure*}[htbp]
  \centering
  \foreach \img/\cap/\lab in {
      mipmap_6_per_arch_lpips_bucket_bpp.png/Performance measured by LPIPS./mipmap_6_lpips,
      mipmap_6_per_arch_vmaf_bucket_bpp.png/Performance measured by VMAF./mipmap_6_vmaf,
      mipmap_6_per_arch_ssim_bucket_bpp.png/Performance measured by SSIM./mipmap_6_ssim,
      mipmap_6_per_arch_psnr_bucket_bpp.png/Performance measured by PSNR./mipmap_6_psnr
  }{%
    \begin{subfigure}{0.45\textwidth}
      \centering
      \includegraphics[width=\linewidth]{figs/\img}
      \caption{\cap} \label{fig:\lab}
    \end{subfigure}
  }
  \caption{Mipmapped texture INR performance against bitrate (bits per pixel).}
  \label{fig:mipmap_6_evals}
\end{figure*}


\subsection{INR-space generation evaluation}
\begin{figure}[htbp]
  \centering
  \includegraphics[width=\linewidth]{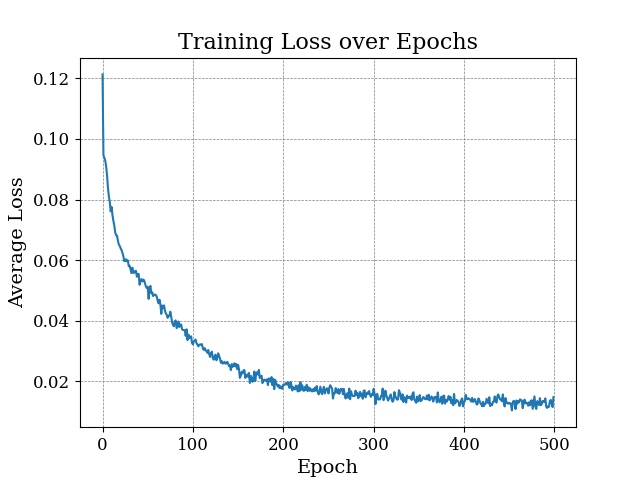}
  \caption{INR-space Diffusion model training loss.}
  \label{fig:diff_train_loss}
\end{figure}

\begin{figure}[htbp]
  \centering
  \foreach \i in {0,...,3}{%
    \begin{subfigure}{0.24\linewidth}
      \centering
      \includegraphics[width=\linewidth]{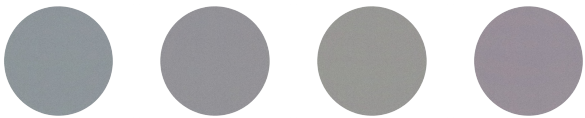}
    \end{subfigure}\hfill
  }
  \caption{Generated texture samples from INR-space diffusion.}
  \label{fig:diff_res}
\end{figure}

We include the INR-space diffusion training loss in Figure \ref{fig:diff_train_loss}, generative samples in Figure \ref{fig:diff_res}.



We argue that the generative quality is yet to be perfect, despite model convergence. This might be limited by scarce training INR weights (around 600). We will explore with increase of training dataset and evaluate the fidelity and coverage between generative and reference set.

%% file: sec/5_conclusions.tex
\section{Conclusions}
\label{sec:conclusions}


In this work, we present how different INRs for textures can be built, with evaluation over their performance, efficiency, memory usage and complexity. In addition, we integrate INR in graphics pipeline, and different down-stream tasks. We have shown that perceptually, INRs can be very effective at image compression, though can struggle on exact details. Despite this, INRs can outperform classical compression methods across all metrics for similar bitrates. We have shown that whilst simple MLPs can achieve good results, Fourier encoding and SIRENs provide significant improvement. We have also shown the importance of hyper-parameter tuning, which can significantly alter the performance for different image types. 

\paragraph{Author contributions.}  Please refer to Section \ref{sec:ack}.


\section{Future directions}  \label{sec:fut-dir}

This paper has demonstrated that for reliable INR compression are consistent process is needed for selecting hyper-parameters. Notably, due to time and computation constraints, we were unable to tune the frequency values used by the SIRENs and Fourier MLPs, which could had significant impacts on the performance. For INR compression to be fast and reliable, an efficient way to select these hyper-parameters should be developed. Future work could explore the selection of hyper-parameters, either through traditional image processing techniques such as the Fourier Transform, or machine learning techniques such as training a model to select the best hyper parameters. 

In addition to simply learning the image's pixels, the model could be trained on a greater variety of randomly sampled points, which may prevent grid-related artefacts when training. 

This paper has also demonstrated the potential of encoding multiple images into one image, in this case through mipmaps without significantly impacting quality. This suggests that even higher compression ratios might be possible when compressing multiple similar images together, which could be used when compressing image libraries, or animations.

The paper's exploration of mipmapping could also be taken further. Effective compression for anisotropic filtering could allow the technique to use less GPU RAM. INRs could also provide novel opportunities in perfectly modelling the filtering, taking in the viewing angle and lod, and outputting the colour averaged over the pixel area. 

In addition to better representing filtering, INRs could also efficiently represent Spatially Varying BiDirectional Scattering Distribution Function (SVBSDF) \cite{M3ashy2026}, allow for more expressive materials to be efficiently stored in memory.

%% file: sec/X_suppl.tex
\clearpage
\setcounter{page}{1}
\maketitlesupplementary

\section{Acknowledgement} \label{sec:ack}

We are deeply grateful to our supervisor, Dounia Hammou, for her insightful guidance and detailed suggestions throughout this project and in the preparation of this report. 

Special thanks to Professor Rafał Mantiuk for lecturing and organizing the Advanced Graphics and Image Processing module, which has been enjoyable along the way.

\paragraph{Author contributions.} Both authors jointly discussed and researched relevant prior work, assisted each other with debugging, and participated in reviewing and improving the overall codebase.

- Albert proposed the idea of adopting SIREN, implemented the multi-resolution hash encoding, expanded the evaluation to cover the entire dataset, and added in evaluation with LPIPS and VMAF. He also proposed the idea of utilising traditional image metrics to select a subset of the Describable Textures Dataset for evaluation. He authored the Evaluation section, including the performance-bitrate plot with confidence intervals, and as an extension, explored the use of INRs for mipmapping.

- Zheyuan (Peter) performed the data analysis and selection, implemented the naïve MLP and its periodic-activated and Fourier-encoded variant, the code for calculating the MAE, MSE, PSNR, and SSIM, integrated with Mitsuba renderer, and evaluated models on individual images. He also chose the images to evaluate in the dataset. He authored the Method section and, as an extension, investigated INR-space generation.

\section{Selected dataset}
\begin{figure}[htbp]
    \centering
    \includegraphics[width=\linewidth]{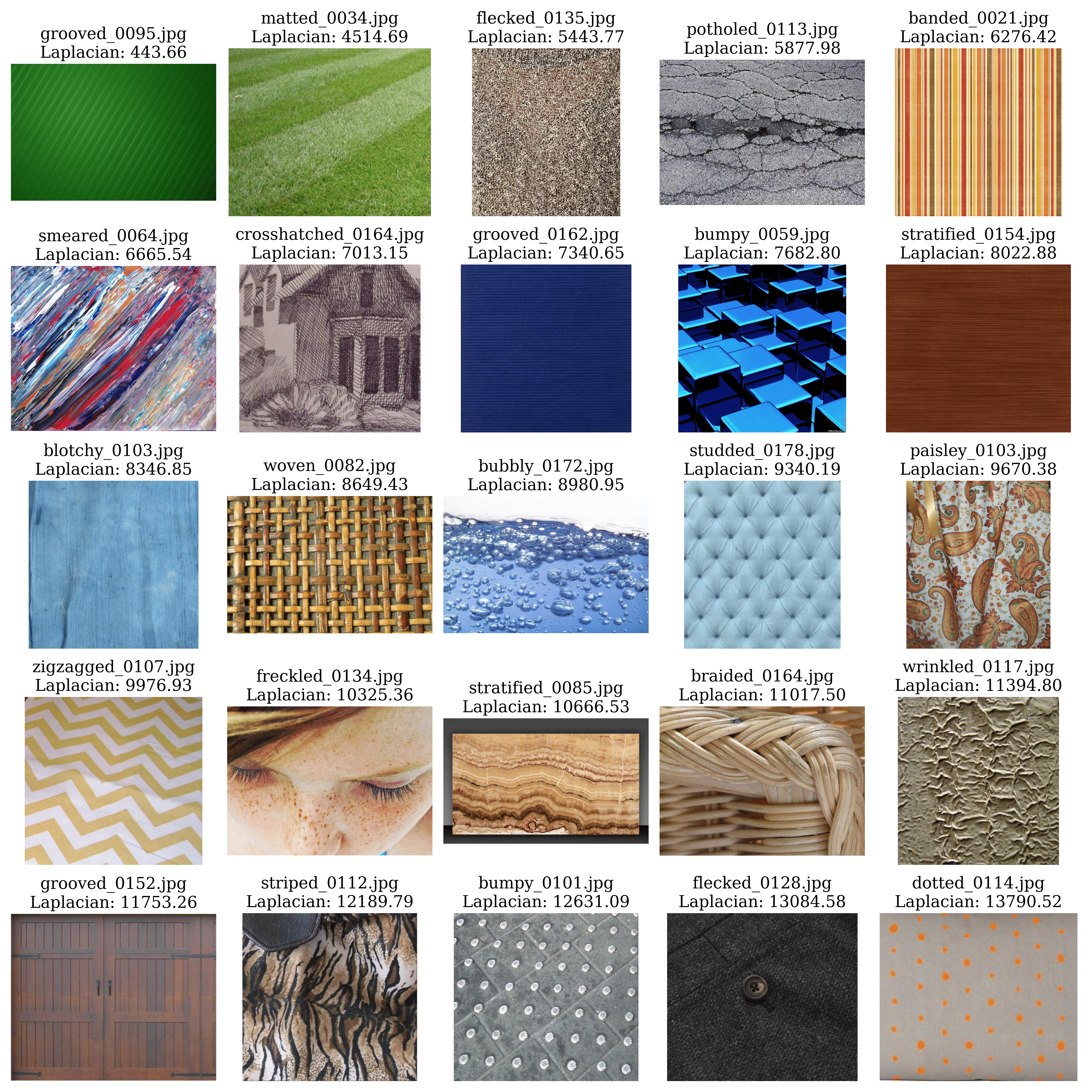}
    \caption{Sampled textures with their Laplacian response.}
    \label{fig:sampled-textures}
\end{figure}

As per \S \ref{sec:method}, we visualize the selected 25 textures in Figure \ref{fig:sampled-textures}.

\section{Methodology} \label{sec:supp-method}



\section{Addition results} 

\subsection{Evaluation of one texture sample} 
\begin{figure}[htbp]
  \centering
  \includegraphics[width=\linewidth]{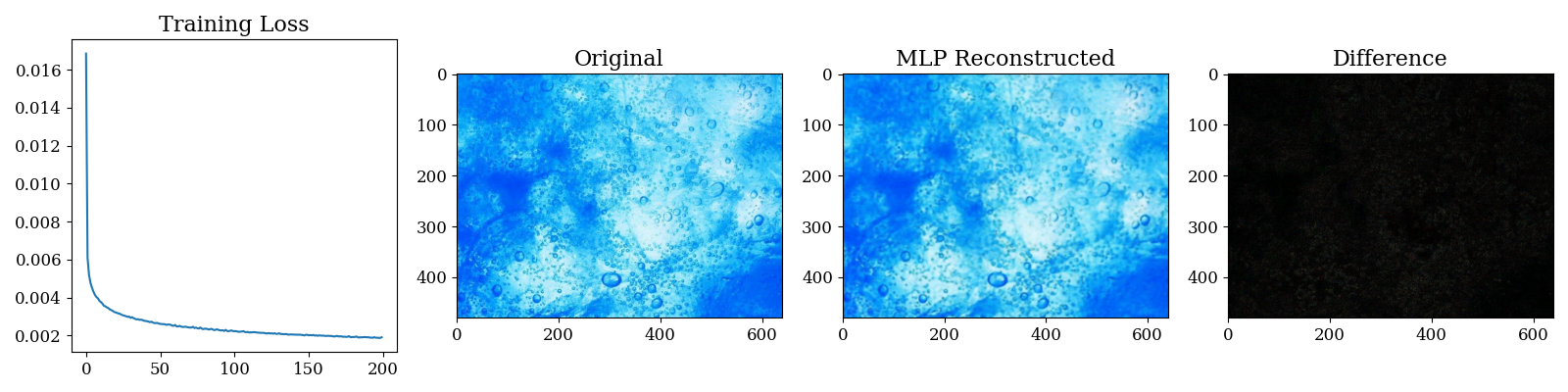}
  \caption{Qualitative visualization comparison.}
  \label{fig:vis-comparison}
\end{figure}

For a single data sample \texttt{bubbly\textunderscore 0122}, we visualize the training loss function, reconstructed texture, and residual error in Figure \ref{fig:vis-comparison}. The error metric is reported in Table \ref{tab:evaluation_metrics}.

\begin{table}[!ht]
\centering
\caption{Evaluation Metrics}
\begin{tabular}{l c}
\hline
\textbf{Metric} & \textbf{Value} \\
\hline
MAE  & 8.0592 \\
MSE  & 121.2885 \\
PSNR (dB) & 27.29 \\
SSIM & 0.9881 \\
\hline
\end{tabular}
\label{tab:evaluation_metrics}
\end{table}



\subsection{Renderer: INR integration} 
\begin{figure}[htbp]
  \centering
  \begin{subfigure}{0.45\linewidth}
    \centering
    \includegraphics[width=\linewidth]{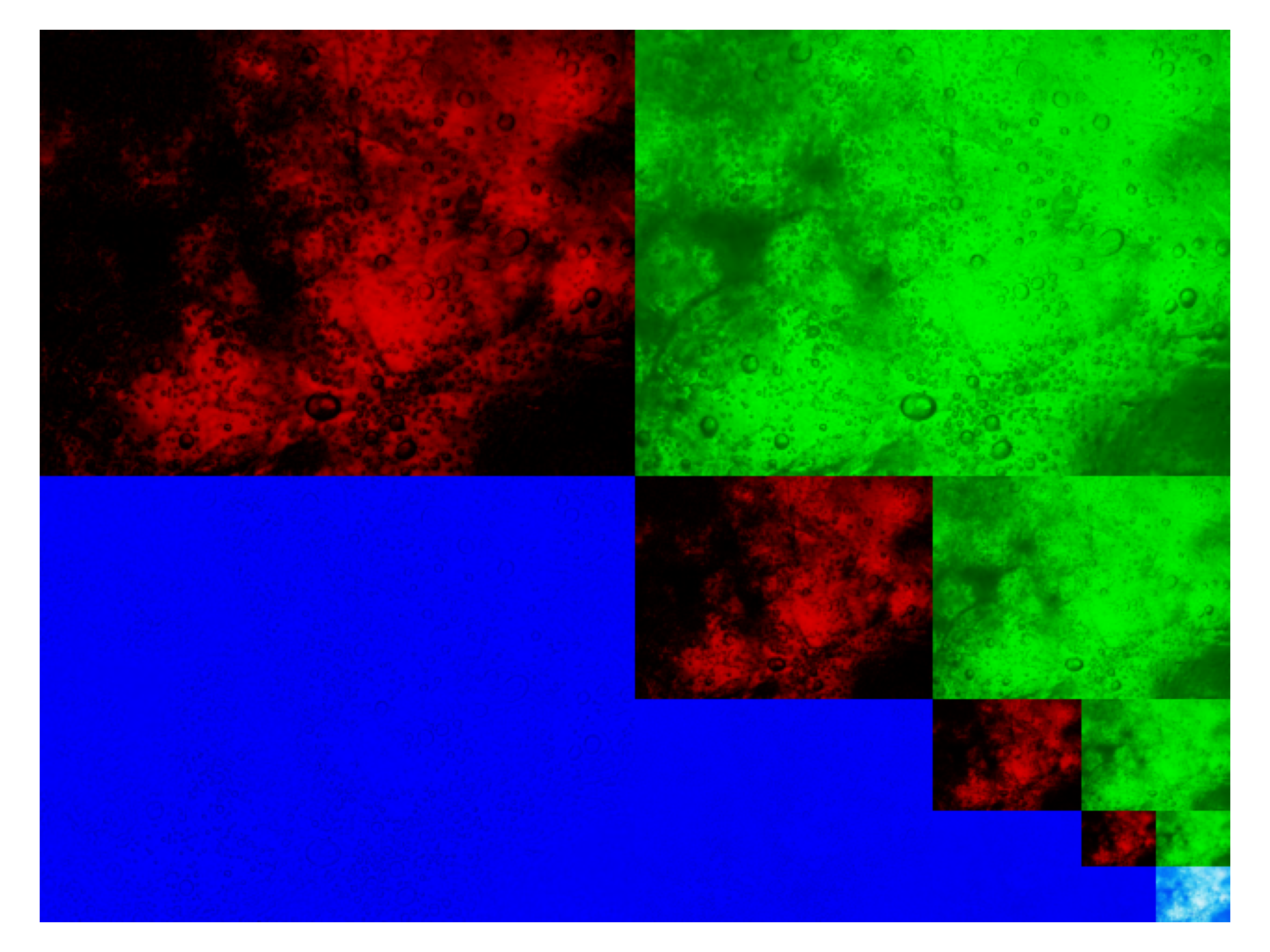}
    \caption{Input image mipmap.}
    \label{fig:renderer-mipmap}
  \end{subfigure}
  \hfill
  \begin{subfigure}{0.45\linewidth}
    \centering
    \includegraphics[width=\linewidth]{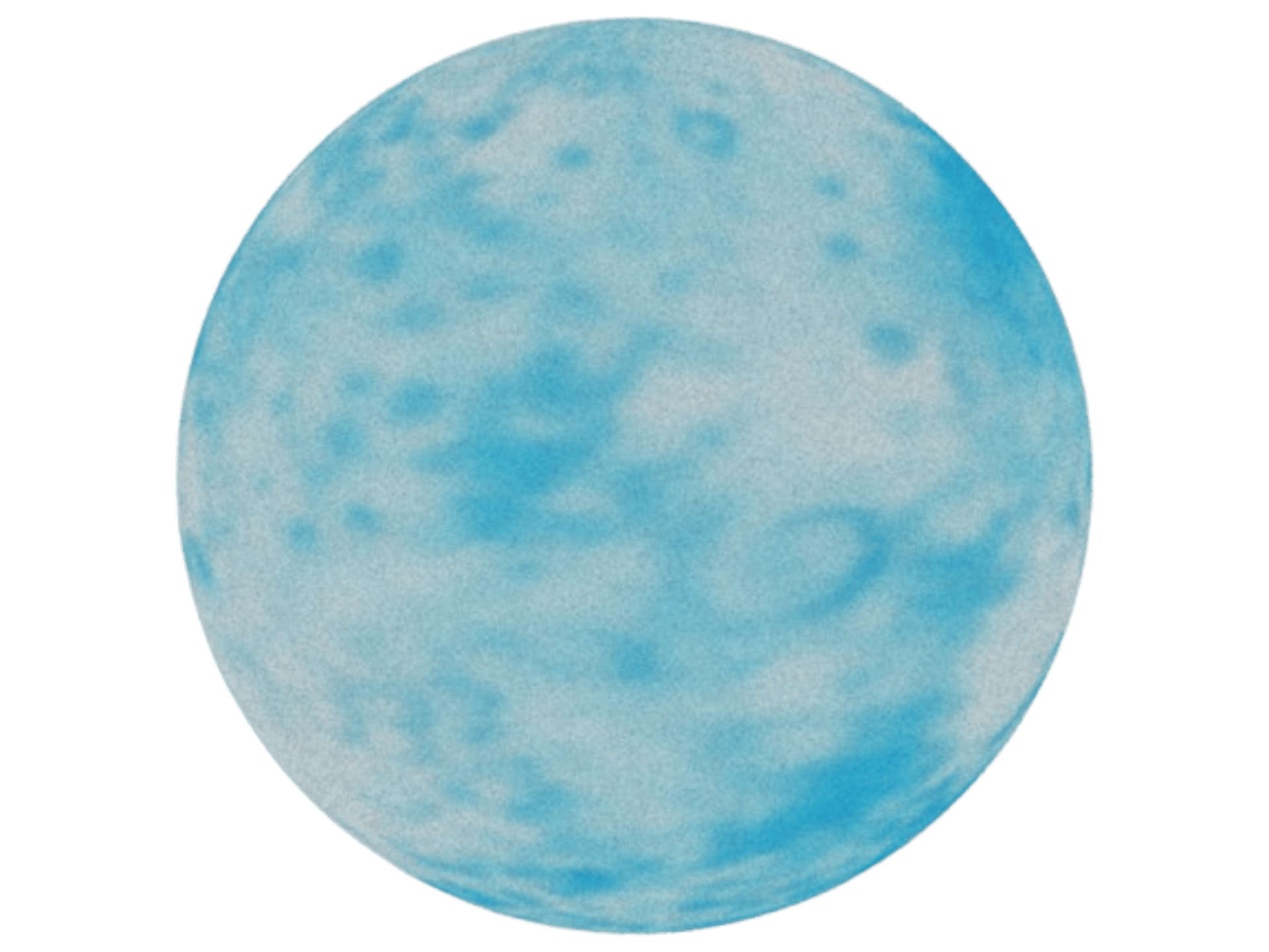}
    \caption{Rendered result from real-time INR plugin.}
    \label{fig:renderer-sphere}
  \end{subfigure}
\end{figure}

We demonstrate the three-colour-channel mipmap pyramid of the texture \texttt{bubbly\textunderscore 0122} in Figure \ref{fig:renderer-mipmap}. With integration with Mitsuba 3 renderer, we are able to render materials real-time from MLP weights. Figure \ref{fig:renderer-sphere} shows the renderer result from INR weights of \texttt{bubbly\textunderscore 0122}.

%% file: bids/project_INR_bid.tex
\section*{Project 1: Implicit neural representation of textures.}

\subsection*{Description (max 250)}

The project area of Implicit Neural Representations (INRs) with periodic activation functions follows a growing interest in using neural networks to directly represent continuous, coordinate-based signals (such as images \cite{Strumpler2022INRImg}, shapes, audio, or physical fields) rather than discrete grids or explicit parametrizations \cite{Sitzmann2020INRActivation}.

A key insight is that standard networks (e.g., multilayer perceptrons with ReLU activations) suffer from a so-called spectral bias (or frequency principle): they tend to fit low-frequency components of the target and struggle with fine high-frequency variation. To solve this, \cite{Sitzmann2020INRActivation} proposes replacing standard activations with sinusoidal (periodic) activations (so-called “SIRENs”) enables the network to represent signals and their derivatives with higher fidelity and expressivity.

The project explores how choosing the architecture, activation function, and initialization for coordinate-based neural networks influences their spectral capacity, smoothness properties, and ability to encode derivatives — and periodic activation functions represent a powerful design choice in this context.

\subsection*{Methodology (max 250)}

The approach in \cite{Sitzmann2020INRActivation} combines architectural choices, input encodings, and training regimes to obtain accurate, stable implicit neural representations that capture high-frequency detail and well-behaved derivatives. A SIREN-style multilayer perceptron will serve as the core model: fully connected layers with sinusoidal activations and the SIREN initialization to allow expressive periodic basis functions. 

As an alternative baseline, models with Fourier feature input mappings have been implemented to evaluate tradeoffs between input encoding and periodic activations.  Optimization uses Adam with learning-rate scheduling, and regularization techniques (weight decay, gradient clipping) prevent instabilities caused by high-frequency modes. 

Evaluation metrics include PSNR/SSIM for reconstruction, spectral error, and L2 norm of differential errors for derivative accuracy. Ablations vary activation types, initialization scales, network depth/width, and input encoding bandwidth to quantify their effects.

\subsection*{Justification (max 150)}

The applicants combine a strong foundation in machine learning and graphics with hands-on experience in neural representation research. Practical skills include the implementation of deep MLP architectures and automatic differentiation for derivative-based losses; proficiency in PyTorch enables rapid prototyping and reproducible experiments. The candidates also have proficiency in OpenGL, enabling the results to be tested with shaders. 

Prior projects have involved material reconstruction. The candidates demonstrate methodological rigour in experimental design and quantitative evaluation, ensuring that findings will be robust, interpretable, and of direct relevance to the INR community.

\subsection*{Work list}
The work list is included, incorporating suggestions provided by Dounia Hammou and Professor Rafał Mantiuk.

* Focus on the benefits of implicit neural representation (a continuous function rather than a discrete set of samples, resolution can be high).

* Generate mipmap levels directly from the neural representation, rather than storing precomputed ones.  Compare with other methods?

* Extension: sampling from such representation - (anisotropic) texture filtering.

* Extension: how to compress multiple similar textures into one neural representation, and report the compression efficiency-quality tradoff.

* Try at least two different implicit representations.

* Extension: build an impressive off-line  rendering system - all can be done in PyTorch.

%% file: main.bib
@inproceedings{Sitzmann2020INRActivation,
author = {Sitzmann, Vincent and Martel, Julien N. P. and Bergman, Alexander W. and Lindell, David B. and Wetzstein, Gordon},
title = {{Implicit neural representations with periodic activation functions}},
year = {2020},
isbn = {9781713829546},
publisher = {Curran Associates Inc.},
address = {Red Hook, NY, USA},
booktitle = {Proceedings of the 34th International Conference on Neural Information Processing Systems},
articleno = {626},
numpages = {12},
location = {Vancouver, BC, Canada},
series = {NIPS '20}
}

@inproceedings{Strumpler2022INRImg,
author = {Str\"{u}mpler, Yannick and Postels, Janis and Yang, Ren and Gool, Luc Van and Tombari, Federico},
title = {{Implicit Neural Representations for Image Compression}},
year = {2022},
isbn = {978-3-031-19808-3},
publisher = {Springer-Verlag},
address = {Berlin, Heidelberg},
url = {https://doi.org/10.1007/978-3-031-19809-0_5},
doi = {10.1007/978-3-031-19809-0_5},
booktitle = {Computer Vision – ECCV 2022: 17th European Conference, Tel Aviv, Israel, October 23–27, 2022, Proceedings, Part XXVI},
pages = {74–91},
numpages = {18},
location = {Tel Aviv, Israel}
}

@software{mitsuba3,
  title = {{Mitsuba 3 renderer}},
  author = {Wenzel Jakob and Sébastien Speierer and Nicolas Roussel and Merlin Nimier-David and Delio Vicini and Tizian Zeltner and Baptiste Nicolet and Miguel Crespo and Vincent Leroy and Ziyi Zhang},
  note = {https://mitsuba-renderer.org},
  version = {3.1.1},
  year = {2022}
}

@InProceedings{cimpoi14DTDTex,
  Author    = {M. Cimpoi and S. Maji and I. Kokkinos and S. Mohamed and A. Vedaldi},
  Title     = {{Describing Textures in the Wild}},
  Booktitle = {Proceedings of the {IEEE} Conf. on Computer Vision and Pattern Recognition ({CVPR})},
  Year      = {2014}}

@inproceedings{Tancik2020FourierEncoding,
author = {Tancik, Matthew and Srinivasan, Pratul P. and Mildenhall, Ben and Fridovich-Keil, Sara and Raghavan, Nithin and Singhal, Utkarsh and Ramamoorthi, Ravi and Barron, Jonathan T. and Ng, Ren},
title = {{Fourier features let networks learn high frequency functions in low dimensional domains}},
year = {2020},
isbn = {9781713829546},
publisher = {Curran Associates Inc.},
address = {Red Hook, NY, USA},
booktitle = {Proceedings of the 34th International Conference on Neural Information Processing Systems},
articleno = {632},
numpages = {11},
location = {Vancouver, BC, Canada},
series = {NIPS '20}
}

@article{Popescu2009MLP,
author = {Popescu, Marius-Constantin and Balas, Valentina E. and Perescu-Popescu, Liliana and Mastorakis, Nikos},
title = {{Multilayer perceptron and neural networks}},
year = {2009},
issue_date = {July 2009},
publisher = {World Scientific and Engineering Academy and Society (WSEAS)},
address = {Stevens Point, Wisconsin, USA},
volume = {8},
number = {7},
issn = {1109-2734},
journal = {WSEAS Trans. Cir. and Sys.},
month = jul,
pages = {579–588},
numpages = {10}
}

@misc{memon2016imagequality,
      title={{Image Quality Assessment for Performance Evaluation of Focus Measure Operators}}, 
      author={Farida Memon and Mukhtiar Ali Unar and Sheeraz Memon},
      year={2016},
      eprint={1604.00546},
      archivePrefix={arXiv},
      primaryClass={cs.CV},
      url={https://arxiv.org/abs/1604.00546}, 
}

@misc{agarap2019ReLU,
      title={{Deep Learning using Rectified Linear Units (ReLU)}}, 
      author={Abien Fred Agarap},
      year={2019},
      eprint={1803.08375},
      archivePrefix={arXiv},
      primaryClass={cs.NE},
      url={https://arxiv.org/abs/1803.08375}, 
}

@article{Muller2022HashEncoding,
author = {M\"{u}ller, Thomas and Evans, Alex and Schied, Christoph and Keller, Alexander},
title = {{Instant neural graphics primitives with a multiresolution hash encoding}},
year = {2022},
issue_date = {July 2022},
publisher = {Association for Computing Machinery},
address = {New York, NY, USA},
volume = {41},
number = {4},
issn = {0730-0301},
url = {https://doi.org/10.1145/3528223.3530127},
doi = {10.1145/3528223.3530127},
journal = {ACM Trans. Graph.},
month = jul,
articleno = {102},
numpages = {15}
}

@article{HORNIK1989MLPUniversal,
title = {{Multilayer feedforward networks are universal approximators}},
journal = {Neural Networks},
volume = {2},
number = {5},
pages = {359-366},
year = {1989},
issn = {0893-6080},
doi = {https://doi.org/10.1016/0893-6080(89)90020-8},
url = {https://www.sciencedirect.com/science/article/pii/0893608089900208},
author = {Kurt Hornik and Maxwell Stinchcombe and Halbert White},
}

@inproceedings{Erkoc_2023_HyperDiffusion,
  author    = {Ziya Erko\c{c} and Fangchang Ma and Qi Shan and Matthias Nie\ss{}ner and Angela Dai},
  title     = {{HyperDiffusion: Generating Implicit Neural Fields with Weight-Space Diffusion}},
  booktitle = {Proceedings of the IEEE/CVF International Conference on Computer Vision (ICCV)},
  year      = {2023},
  pages     = {14300--14310},
  url       = {https://ziyaerkoc.com/hyperdiffusion},
  organization = {IEEE/CVF}
}

@INPROCEEDINGS{Hore2010PSNR,
  author={Horé, Alain and Ziou, Djemel},
  booktitle={2010 20th International Conference on Pattern Recognition}, 
  title={{Image Quality Metrics: PSNR vs. SSIM}}, 
  year={2010},
  volume={},
  number={},
  pages={2366-2369},
  doi={10.1109/ICPR.2010.579}}

@ARTICLE{Wang2004SSIM,
  author={Zhou Wang and Bovik, A.C. and Sheikh, H.R. and Simoncelli, E.P.},
  journal={IEEE Transactions on Image Processing}, 
  title={{Image quality assessment: from error visibility to structural similarity}}, 
  year={2004},
  volume={13},
  number={4},
  pages={600-612},
  doi={10.1109/TIP.2003.819861}}

@inproceedings{zhang2018perceptual,
  title={{The Unreasonable Effectiveness of Deep Features as a Perceptual Metric}},
  author={Zhang, Richard and Isola, Phillip and Efros, Alexei A and Shechtman, Eli and Wang, Oliver},
  booktitle={CVPR},
  year={2018}
}

@INPROCEEDINGS{Silpa08TexMemoryUsage,
  author={Silpa, B. V. N. and Patney, Anjul and Krishna, Tushar and Panda, Preeti Ranjan and Visweswaran, G. S.},
  booktitle={2008 IEEE/ACM International Conference on Computer-Aided Design}, 
  title={Texture Filter Memory — a power-efficient and scalable texture memory architecture for mobile graphics processors}, 
  year={2008},
  volume={},
  number={},
  pages={559-564},
  doi={10.1109/ICCAD.2008.4681631}}

@inproceedings{
    M3ashy2026, 
    author = {Chenliang Zhou and Zheyuan Hu and Alejandro Sztrajman and Yancheng Cai and Yaru Liu and Cengiz Oztireli}, 
    title = {M$^{3}$ashy: Multi-Modal Material Synthesis via Hyperdiffusion}, 
    year = {2026}, 
    booktitle = {Proceedings of the 40th AAAI Conference on Artificial Intelligence}, 
    location = {Singapore}, 
    series = {AAAI'26} 
}

@misc{su2025chord,
      title={CHOrD: Generation of Collision-Free, House-Scale, and Organized Digital Twins for 3D Indoor Scenes with Controllable Floor Plans and Optimal Layouts}, 
      author={Chong Su and Yingbin Fu and Zheyuan Hu and Jing Yang and Param Hanji and Shaojun Wang and Xuan Zhao and Cengiz Öztireli and Fangcheng Zhong},
      year={2025},
      eprint={2503.11958},
      archivePrefix={arXiv},
      primaryClass={cs.CV},
      url={https://arxiv.org/abs/2503.11958}, 
}

@inproceedings{Nystad2012ASTC,
author = {Nystad, J. and Lassen, A. and Pomianowski, A. and Ellis, S. and Olson, T.},
title = {Adaptive scalable texture compression},
year = {2012},
isbn = {9783905674415},
publisher = {Eurographics Association},
address = {Goslar, DEU},
booktitle = {Proceedings of the Fourth ACM SIGGRAPH / Eurographics Conference on High-Performance Graphics},
pages = {105–114},
numpages = {10},
location = {Paris, France},
series = {EGGH-HPG'12}
}

@techreport{JPEG,
  author       = {{CCITT}},
  title        = {T.81 -- Digital Compression and Coding of Continuous-Tone Still Images -- Requirements and Guidelines},
  institution  = {International Telegraph and Telephone Consultative Committee},
  year         = {1992},
  month        = sep,
  url          = {https://www.w3.org/Graphics/JPEG/itu-t81.pdf},
  note         = {Accessed: 2025-11-11}
}

@inproceedings{li2016vmaf,
  title={{VMAF}: The Journey Continues},
  author={Li, Zhi and Aaron, Anne and Katsavounidis, Ioannis and Moorthy, Anush and Manohara, Megha},
  booktitle={The Netflix Tech Blog},
  year={2016},
  note={https://netflixtechblog.com/vmaf-the-journey-continues-44b51ee9ed12}
}

@misc{vmaf-torch,
      title={{VMAF} Re-implementation on {PyTorch}: Some Experimental Results}, 
      author={Kirill Aistov and Maxim Koroteev},
      year={2023},
      eprint={2310.15578},
      archivePrefix={arXiv},
      primaryClass={cs.LG},
        howpublished={https://github.com/alvitrioliks/VMAF-torch}
}
